\documentclass[sigconf]{acmart}

\AtBeginDocument{%
  }

\copyrightyear{2023}
\acmYear{2023}
\setcopyright{rightsretained}
\acmConference[ICAIL 2023]{Nineteenth International Conference on Artificial Intelligence and Law}{June 19--23, 2023}{Braga, Portugal}
\acmBooktitle{Nineteenth International Conference on Artificial Intelligence and Law (ICAIL 2023), June 19--23, 2023, Braga, Portugal}\acmDOI{10.1145/3594536.3595161}
\acmISBN{979-8-4007-0197-9/23/06}

\usepackage{balance}
\usepackage{color,colortbl,soul}
\definecolor{bert}{rgb}{0.796, 0.835, 0.91}
\definecolor{rf}{rgb}{0.7, 0.89, 0.8}

\usepackage{fancyvrb}
\usepackage[relative,overlay]{textpos}
\usepackage{tikz}
\usepackage{relsize}

\newcommand*\circled[1]{\tikz[baseline=(char.base)]{
    \node[shape=circle,fill=black, text=white,draw,inner sep=2pt] (char) {#1};}}

\makeatletter
\gdef\@copyrightpermission{
  \begin{minipage}{0.3\columnwidth}
   \href{https://creativecommons.org/licenses/by/4.0/}{\includegraphics[width=0.90\textwidth]{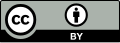}}
  \end{minipage}\hfill
  \begin{minipage}{0.7\columnwidth}
   \href{https://creativecommons.org/licenses/by/4.0/}{This work is licensed under a Creative Commons Attribution International 4.0 License.}
  \end{minipage}
  \vspace{5pt}
}
\makeatother

\begin{document}

\title[Evaluation of GPT for Semantic Annotation of Legal Texts]{Unlocking Practical Applications in Legal Domain: Evaluation of GPT for Zero-Shot Semantic Annotation of Legal Texts}

\author{Jaromir Savelka}
\affiliation{%
  \institution{Carnegie Mellon University}
  \city{Pittsburgh}
  \state{PA}
  \country{USA}
  \postcode{15221}
}
\email{jsavelka@cs.cmu.edu}

\renewcommand{\shortauthors}{Savelka}

\begin{abstract}
We evaluated the capability of a state-of-the-art generative pre-trained transformer~(GPT) model to perform semantic annotation of short text snippets~(one to few sentences) coming from legal documents of various types. Discussions of potential uses~(e.g.,~document drafting, summarization) of this emerging technology in legal domain have intensified, but to date there has not been a rigorous analysis of these large language models' (LLM) capacity in sentence-level semantic annotation of legal texts in zero-shot learning settings. Yet, this particular type of use could unlock many practical applications~(e.g.,~in contract review) and research opportunities~(e.g.,~in empirical legal studies). We fill the gap with this study.  We examined if and how successfully the model can semantically annotate small batches of short text snippets (10--50) based exclusively on concise definitions of the semantic types. We found that the GPT model performs surprisingly well in zero-shot settings on diverse types of documents (F$_1=.73$ on a task involving court opinions, $.86$ for contracts, and $.54$ for statutes and regulations). These findings can be leveraged by legal scholars and practicing lawyers alike to guide their decisions in integrating LLMs in wide range of workflows involving semantic annotation of legal texts.
\end{abstract}

\ccsdesc[500]{Applied computing~Law}
\ccsdesc[500]{Applied computing~Annotation}
\keywords{Semantic legal annotation, generative pre-trained transformers, GPT, transfer learning, zero-shot, adjudicatory decisions, contracts, statutory and regulatory provisions}
\maketitle

\section{Introduction}

We evaluate the effectiveness of GPT-3.5 (\verb|text-davinci-003|) in tasks focused on (i) contract review, (ii) statutory/regulatory provisions investigation and (iii) case-law analysis. We benchmark the performance of the general (not fine-tuned) GPT-3.5 model in performing annotation of small batches of short text snippets coming from the aforementioned types of legal documents against the performance of a traditional statistical machine learning (ML) model (random forest) and fine-tuned BERT model (RoBERTa). The GPT moodel's annotations are based on compact one sentence long semantic type definitions provided to the model as a prompt. Specifically, we analyzed the following research question in the context of the three legal annotation tasks: Given brief type definitions from a single non-hierarchical type system describing short snippets of text, how successfully can a general GPT-3.5 model automatically classify such text spans in terms of the type system's categories?

\section{Related Work}
\textbf{Zero-shot GPT in AI \& Law}. Yu et al. applied GPT to the COLIEE entailment task, improving on the then existing state-of-the-art~\cite{https://doi.org/10.48550/arxiv.2212.01326}. Bommarito and Katz successfully applied GPT-3.5 and GPT-4 to the Bar Examination~\cite{bommarito2022gpt,katz2023gpt}. Other use cases include assessment of trademark distinctiveness~\cite{goodhue2023classification}, legal reasoning~\cite{blair2023can,nguyen2023well}, and U.S. Supreme court judgment modeling~\cite{hamilton2023blind}.

\textbf{Rhetorical/Functional Segments in Adjudicatory Decisions.} The task involves labeling of smaller textual snippets such as sentences~\cite{savelka2017sentence} in terms of, e.g., rhetorical roles, functional or argument units. Examples include court~\cite{savelka2017using} or administrative decisions from the U.S.~\cite{walker2019automatic}, multi-domain court decisions from India~\cite{bhattacharya2019identification} or Canada~\cite{xu2021accounting,xu2021toward}, international court~\cite{poudyal2020echr} or arbitration decisions~\cite{branting2019semi}, or even multi-\{domain,country\} adjudicatory decisions~\cite{savelka2020cross}. Identifying a section that states an outcome of the case has also received considerable attention separately~\cite{xu2020using,petrova2020extracting}. The task sometimes takes a form of identifying a small number of contiguous parts typically comprising multiple paragraphs. Different variations of this task were applied to several legal domains from countries such as Canada~\cite{farzindar2004letsum}, the Czech Republic~\cite{harasta2019automatic}, France~\cite{boniolperformance}, the U.S.~\cite{savelka2018segmenting}, or even in multi-jurisdictional settings~\cite{savelka2021lex}.

\textbf{Classification of Legal Norms.} Researchers used traditional statistical supervised ML models to classify portions of Italian statutory texts with types, such as definition, prohibition, or obligation~\cite{francesconi2010integrating,biagioli2005automatic}. Other groups classified sentences from Dutch statutory texts in terms of categories such as definition, publication provision, or scope of change~\cite{de2010machine}. Some work focuses on fine-grained semantic analysis of statutory texts in terms of obligations, permissions, subject agents or themes~\cite{wyner2011rule,vsavelka2015applying,vsavelka2015transfer}, concepts or definitions~\cite{winkelsm2012automatic}.

\textbf{Classification of contractual clauses.}. Chalkidis et al. analyzed contractual clauses in terms of types such as termination clause, governing law or jurisdiction~\cite{chalkidis2017extracting,chalkidis2021neural}. Leivaditi et al. released a benchmark data set of 179 lease agreement documents focusing on recognition of entities and red flags~\cite{leivaditi2020benchmark}. In this work we focus on twelve selected semantic types from the Contract Understanding Atticus Dataset (CUAD)~\cite{hendrycks2021cuad}. Wang et al. assembled and released the Merger Agreement Understanding Dataset (MAUD)~\cite{wang2023maud}.

\section{Data}
We use three existing manually annotated data sets. Each  supports various tasks involving different types of legal documents. All of them are equipped with expert annotations attached to (usually) short pieces of text. We further filtered and processed the data sets to make them suitable for this work's experiments.

\begin{figure}[t]
    \raisebox{14.5pt}{\includegraphics[height=4.15cm]{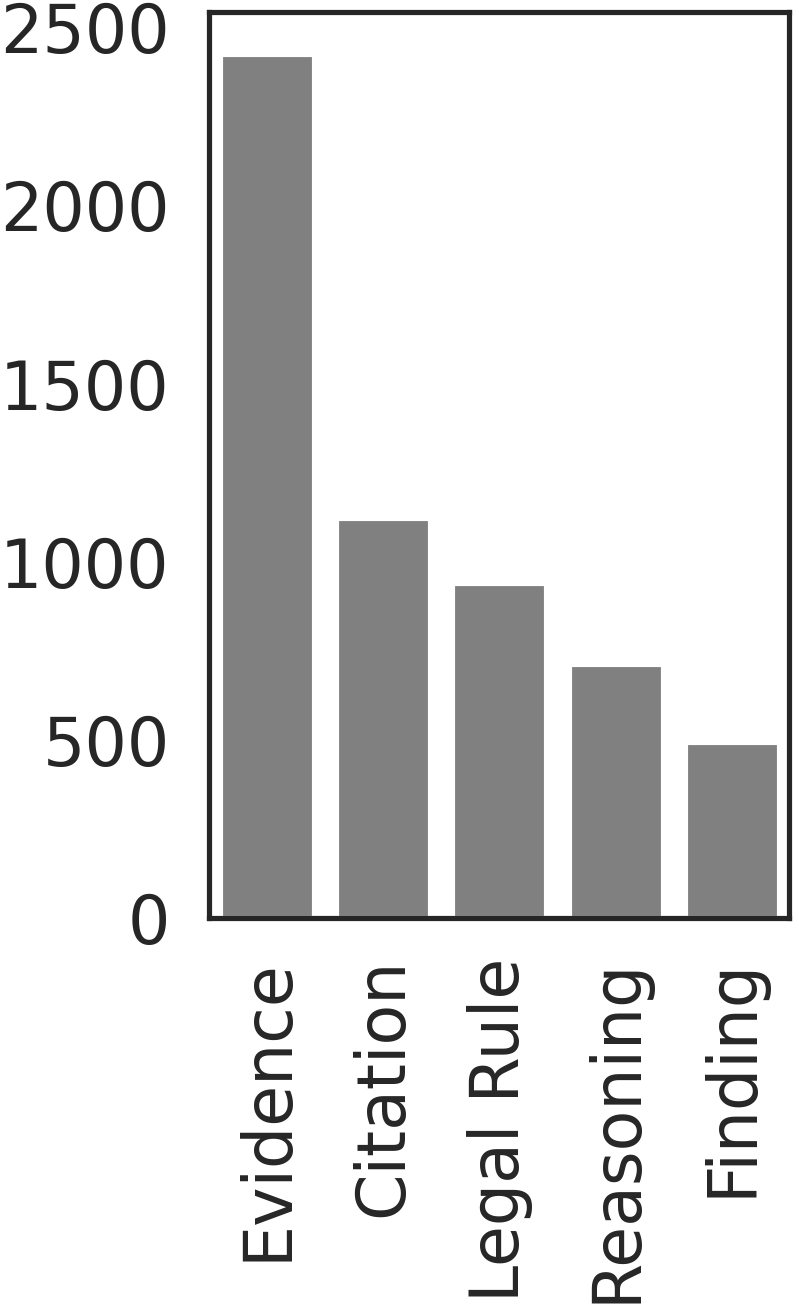}}
    \includegraphics[height=4.7cm]{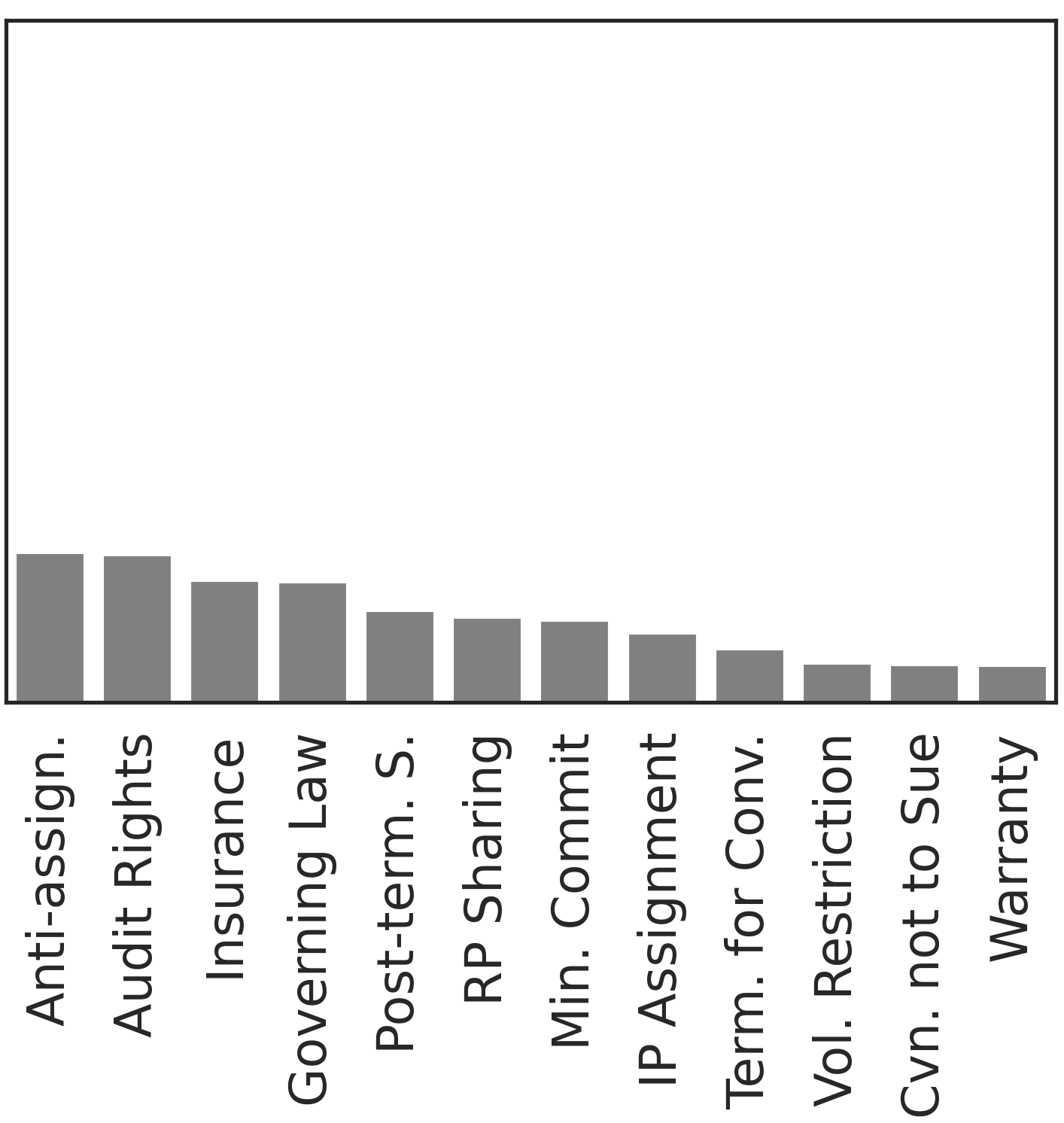}
    \raisebox{6.5pt}{\includegraphics[height=4.5cm]{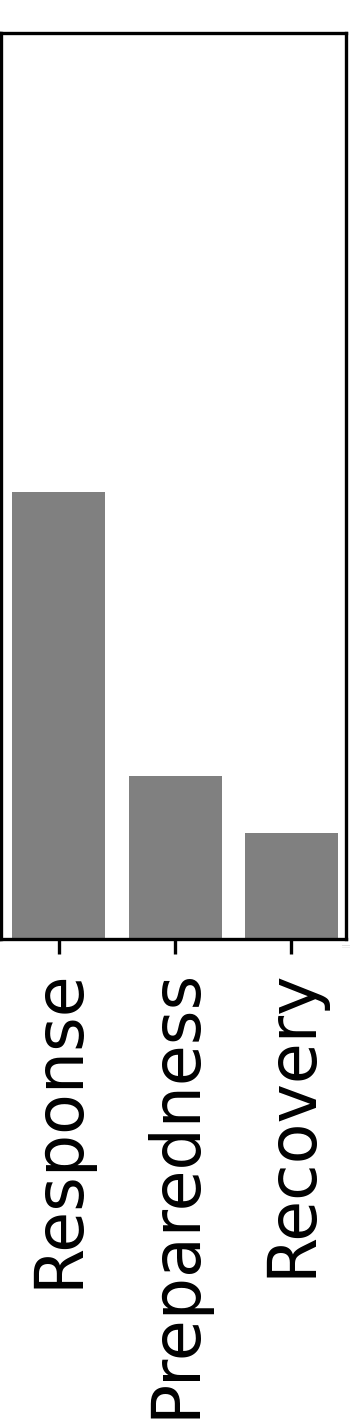}}
\begin{textblock*}{3.4cm}(0.6cm,-4.55cm)
\Small
BVA
\end{textblock*}
\begin{textblock*}{3.4cm}(6.15cm,-4.55cm)
\Small
PHASYS
\end{textblock*}
\begin{textblock*}{3.4cm}(5.05cm,-4.55cm)
\Small
CUAD
\end{textblock*}
    \caption{Semantic Types Distribution. The figure shows distribution of the semantic types across the three data sets.}
    \label{fig:data_dist}
\end{figure}

The U.S. Board of Veterans' Appeals\footnote{U.S. Department of Veterans' Appeals: Board of Veteran's Appeals. Available at: \url{https://www.bva.va.gov/} [Accessed 2023-02-09]} (BVA) is an administrative body within the U.S. Department of Veterans Affairs (VA) responsible for hearing appeals from veterans who are dissatisfied with decisions made by VA regional offices. The BVA reviews a wide range of issues, including claims for disability compensation, survivor benefits, and other compensation and pension claims. Walker et al.~\cite{walker2019automatic} analyzed 50 BVA decisions issued between 2013 and 2017. The decisions were all arbitrarily selected cases dealing with claims by veterans for service-related post-traumatic stress disorder (PTSD). For each decision, the researchers manually extracted sentences addressing the factual issues. The sentences were then manually annotated with rhetorical roles they play in the respective decisions~\cite{walker2017semantic}. Figure \ref{fig:data_dist} (left) shows the distribution of the labels.

Contract Understanding Atticus Dataset (CUAD) is a corpus of 510 commercial legal contracts that have been manually labeled under the supervision of professional lawyers. This effort resulted in more than 13,000 annotations.\footnote{The Atticus Project: Contract Understanding Atticus Dataset (CUAD). Available at: \url{https://www.atticusprojectai.org/cuad} [Accessed 2023-02-09]} The data set was released by Hendrycks et al.~\cite{hendrycks2021cuad} and it identifies 41 types of legal clauses that are typically considered important in contract review in connection with corporate transactions.  In this study, we decided to work with the 12 most common clause-level types present in the corpus the distribution of which is shown in Figure \ref{fig:data_dist} (center).

At the University of Pittsburgh's Graduate School of Public Health, researchers have manually coded federal, state and local laws and regulations related to emergency preparedness and response of the public health system (PHS). They used the codes to analyze network diagrams representing various functional features of states' regulatory frameworks for public health emergency preparedness. They retrieved candidate sets of relevant statutes and regulations from a full-text legal information service and identified relevant spans of text~\cite{sweeney2013network}. They then coded the relevant spans as per instructions in the codebook,\footnote{PHASYS ARM 2 – LEIP Code Book. Available at: \url{https://www.phasys.pitt.edu/pdf/Code_Book_Numerical_Defintions.pdf} [Accessed 2023-02-09]} representing relevant features of those spans. In this work we focus on the purpose of the legal provision in terms of the three categories the distribution of which is shown in Figure \ref{fig:data_dist} (right). The statutory and regulatory texts were automatically divided into text units which are often non-contiguous spans of text referenceable with citations~\cite{savelka2014mining}.

\section{Experiments}

We use the Jaccard similarity measure as the baseline~(tokens as sets). Each text snippet is compared to the type definitions available in the respective task. It is then assigned the label the definition of which has the highest similarity score with the snippet.

\begin{figure}
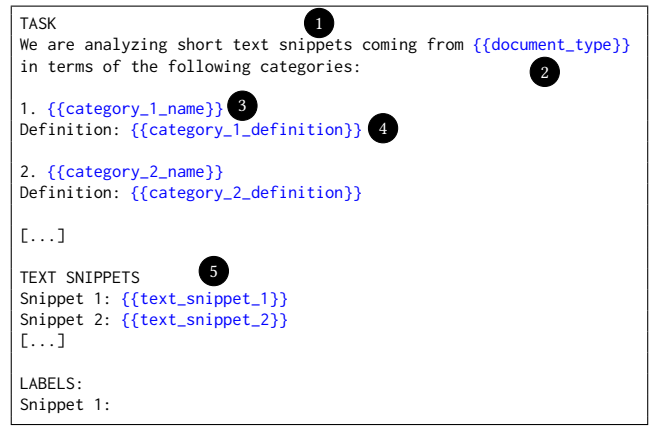

\footnotesize
\begin{Verbatim}[frame=single,commandchars=\\\{\}]
TASK
We are analyzing short text snippets coming from \textcolor{blue}{\string{\string{document_type\string}\string}} 
in terms of the following categories:

1. \textcolor{blue}{\string{\string{category_1_name\string}\string}}
Definition: \textcolor{blue}{\string{\string{category_1_definition\string}\string}}

2. \textcolor{blue}{\string{\string{category_2_name\string}\string}}
Definition: \textcolor{blue}{\string{\string{category_2_definition\string}\string}}

[...]

TEXT SNIPPETS
Snippet 1: \textcolor{blue}{\string{\string{text_snippet_1\string}\string}}
Snippet 2: \textcolor{blue}{\string{\string{text_snippet_2\string}\string}}
[...]

LABELS:
Snippet 1:
\end{Verbatim}
\begin{textblock*}{3.4cm}(2.4cm,-5.6cm)
\circled{1}
\end{textblock*}
\begin{textblock*}{3.4cm}(5.4cm,-4.95cm)
\circled{2}
\end{textblock*}
\begin{textblock*}{3.4cm}(1.38cm,-4.5cm)
\circled{3}
\end{textblock*}
\begin{textblock*}{3.4cm}(3.25cm,-4.2cm)
\circled{4}
\end{textblock*}
\begin{textblock*}{3.4cm}(1cm,-2.3cm)
\circled{5}
\end{textblock*}
\caption{Batch Prediction Prompt Template. The preamble~(1) primes the model to generate semantic type predictions. The tokens 
 surrounded by curly braces are replaced with the document type (2) according to the data set, the names of the semantic types (3), the corresponding definitions (4), and the analyzed text snippets (5).}
\label{fig:prompt-template}
\end{figure}

We benchmark the performance of GPT-3.5~\cite{brown2020language,ouyangtraining} against a traditional statistical supervised learning algorithm~(random forest~\cite{ho1995random}) which is evaluated using 10-fold cross-validation at the document level. Within each iteration of the cross-validation, we utilize grid search\footnote{Scikit learn: GridSearchCV. Available at: \url{https://scikit-learn.org/stable/modules/generated/sklearn.model_selection.GridSearchCV.html} [Accessed 2023-02-05]} to select the best set of hyperparameters. The space that is being considered is defined over the type of n-grams to be used, number of estimators and maximum tree depth.

\begin{table*}[t]
  \caption{Experimental results in terms of micro-F$_1$ scores. The Jaccard column shows the performance of the Jaccard similarity baseline. The @N labels denote how many data points were used in the training of the two supervised ML systems (RandF -- random forest, BERT -- base RoBERTa), where @Max means all the available data points were used. The GPT section reports the performance of the text-davinci-003 model. The blue cells signify the point at which the BERT system matched the performance of either one of the GPT-3.5 models. The green shaded cells do the same for the random forest.}
  \label{tab:results}
  \setlength{\tabcolsep}{2pt}
  \begin{tabular}{l|r|rrrrrrrrrrrrrr|rr}
    \toprule
           & &\multicolumn{2}{c}{@20}&\multicolumn{2}{c}{@50}&\multicolumn{2}{c}{@100}&\multicolumn{2}{c}{@250}&\multicolumn{2}{c}{@500}&\multicolumn{2}{c}{@1000}&\multicolumn{2}{c|}{@Max}&\\
                         &Jaccard& RandF & BERT  & RandF & BERT  & RandF & BERT  & RandF & BERT  & RandF & BERT  & RandF & BERT  & RandF & BERT  & GPT \\
    \midrule
    \bf BVA              &\bf .35&\bf .30&\bf .27&\bf .31&\bf .26&\bf .51&\bf .52&\bf .65&\bf\cellcolor{bert}.82&\bf .65&\bf .85&\bf\cellcolor{rf}.73&\bf .89&\bf .83&\bf .92&\bf .73\\
    Citation             &    .71&    .11&    .00&    .12&    .00&    .77&    .49&    .94&\cellcolor{bert}.97&\cellcolor{rf}.97&    .99&    .97&    1.0&    .99&    1.0&    .96\\
    Evidence             &    .28&    .58&    .59&    .61&    .60&    .69&    .66&    .77&\cellcolor{bert}.91&    .76&    .92&\cellcolor{rf}.81&    .93&    .87&    .94&    .77&\\
    Finding              &    .08&    .22&    .10&    .14&    .00&    .35&    .21&    .50&\cellcolor{bert}.55&    .48&    .65&\cellcolor{rf}.63&    .77&    .74&    .86&    .53&\\
    Legal Rule           &    .15&    .00&    .00&    .06&    .00&    .15&    .15&    .38&\cellcolor{bert}.77&    .41&    .84&    .68&    .91&\cellcolor{rf}.89&    .95&    .72&\\
    Reasoning            &    .24&    .11&    .10&    .04&    .00&    .07&    .00&    .23&\cellcolor{bert}.52&    .15&    .56&    .21&    .64&    .40&    .71&    .43&\\
    \midrule
    \bf CUAD             &\bf .38&\bf .25&\bf .12&\bf .31&\bf .14&\bf .52&\bf .43&\bf .73&\bf .68&\bf .80&\bf\cellcolor{bert}.87&\cellcolor{rf}\bf .86&\bf .93&\bf.89&\bf .95&\bf .86\\
    Anti-assignment      &    .56&    .12&    .00&    .16&    .10&    .78&    .71&    .87&    .81&    .90&    .94&    .93&    .97&    .94&\cellcolor{bert}.99&    .96\\
    Audit Rights         &    .50&    .39&    .12&    .35&    .06&    .49&    .71&    .71&    .82&    .81&\cellcolor{bert}.94&    .86&    .97&    .90&    .96&    .93\\
    Cvnt. not to Sue     &    .31&    .04&    .00&    .00&    .00&    .04&    .00&    .73&    .26&    .76&\cellcolor{bert}.87&    .76&    .93&\cellcolor{rf}.90&    .96&    .78\\
    Governing Law        &    .52&    .80&    .17&    .80&    .22&    .96&    .92&    .94&\cellcolor{bert}1.0&    .98&    1.0&\cellcolor{rf}.99&    1.0&    1.0&    1.0&    .99\\
    IP Assignment        &    .05&    .40&    .00&    .51&    .00&    .53&    .08&    .73&    .13&    .76&\cellcolor{bert}.84&    .83&    .88&\cellcolor{rf}.87&    .94&   .84\\
    Insurance            &    .38&    .11&    .13&    .23&    .13&    .64&    .73&    .93&    .92&    .90&\cellcolor{bert}.97&    .93&    .98&\cellcolor{rf}.95&    .97&    .96\\
    Min. Commitment      &    .20&    .00&    .00&    .23&    .00&    .52&    .00&\cellcolor{rf}.64&    .38&    .72&\cellcolor{bert}.69&    .77&    .87&    .81&    .91&    .67\\
    Post-term. Services  &    .43&    .21&    .15&    .29&    .10&    .44&    .29&    .60&    .57&    .72&\cellcolor{bert}.77&\cellcolor{rf}.77&    .84&    .78&    .85&    .69\\
    Profit Sharing       &    .25&    .08&    .17&    .04&    .16&    .03&    .01&    .56&    .75&    .76&\cellcolor{bert}.88&\cellcolor{rf}.86&    .92&    .87&    .94&    .78\\
    Termination Cnv.     &    .43&    .27&    .00&    .60&    .00&    .42&    .00&    .81&    .81&\cellcolor{rf}.88&\cellcolor{bert}.91&    .88&    .95&    .91&    .97&    .89\\
    Volume Restriction   &    .07&    .00&    .00&    .01&    .00&    .03&    .00&    .08&    .30&    .15&    .41&\cellcolor{rf}.45&\cellcolor{bert}.78&    .61&    .90&    .46\\
    Warranty Duration    &    .21&    .02&    .00&    .12&    .00&    .15&    .00&    .48&    .21&    .56&    .76&    .79&\cellcolor{bert}.91&\cellcolor{rf}.84&    .93&    .81\\
    \midrule
    \bf PHASYS           &\bf .24&\bf .52&\bf .48&\bf .53&\bf .48&\bf .53&\bf .49&\bf\cellcolor{rf}.56&\bf\cellcolor{bert}.65&\bf .59&\bf .68&\bf .61&\bf .72&\bf .63&\bf .72&\bf .54\\
    Preparedness         &    .24&    .01&    .00&    .00&    .00&    .00&    .00&    .08&    .31&    .22&\cellcolor{bert}.46&    .20&    .56&    .31&    .61&    .45\\
    Response             &    .24&\cellcolor{rf}.77&\cellcolor{bert}.77&    .77&    .77&    .77&    .77&    .78&    .79&    .78&    .78&    .79&    .80&    .80&    .81&    .63\\
    Recovery             &    .27&    .27&    .00&\cellcolor{rf}.33&    .00&    .33&    .03&    .38&\cellcolor{bert}.55&    .35&    .59&    .48&    .60&    .53&    .67&    .28\\
    \bottomrule
  \end{tabular}
\end{table*}

To compare the performance of the zero-shot GPT-3.5 model against a fine-tuned LLM, we fine-tune the base RoBERTa model~\cite{Liu2019} for 10 epochs on the training set within each of the cross-validation folds. The same splits as for evaluating the performance of the random forest are used. We set the batch size to 16 and the length of the sequence to 512. As optimizer we use the Adam algorithm~\cite{kingma2014adam} with initial learning rate set to $4e^{-5}$. 

To gauge how many labeled documents are needed by the supervised ML system to match and exceed the performance of GPT-3.5, we train the random forest and RoBERTa on training sets of varying sizes. We train the systems on the training sets with 20, 50, 100, 250, 500 and 1,000 data points. These documents are randomly sampled from the training set in each iteration of the cross-validation.

To test the performance of \verb|text-davinci-003|, we submit a batch of text snippets using the \verb|openai| Python library\footnote{GitHub: OpenAI Python Library. Available at: \url{https://github.com/openai/openai-python} [Accessed 2023-02-09]} which is a wrapper for the OpenAI's REST API. We make the batches as large as possible to achieve maximum cost effectiveness. Their size is determined by the size of the evaluated text snippets that can fit into the prompt (4,097 tokens), leaving enough space for the completion~(i.e., the predictions). For the BVA decisions' sentences the batch size was set to 50, for the CUAD's contractual clauses to 20, and for PHASYS' statutory and regulatory provisions to 10.

We embed each batch in the prompt template shown in Figure \ref{fig:prompt-template}. The model returns the list of predicted labels as the prompt completion. The construction of the prompt is focused on maximizing the cost effectiveness of the proposed approach which may somewhat limit the performance of the evaluated GPT-3.5 model.

We set the \verb|temperature| to 0.0, which corresponds to no randomness. The higher the \verb|temperature| the more creative the output but it can also be less factual. We set \verb|max_tokens| to 500 (a token roughly corresponds to a word). This parameter controls the maximum length of the output. We set \verb|top_p| to 1, as is recommended  when \verb|temperature| is set to 0.0. This parameter is related to \verb|temperature| and also influences creativeness of the output.  We set \verb|frequency_penalty| to 0, which allows repetition by ensuring no penalty is applied to repetitions. Finally, we set \verb|presence_penalty| to 0, ensuring no penalty is applied to tokens appearing multiple times in the output, which is especially important for our use case.

\section{Results and Discussion}

\begin{table*}[t]
  \caption{Confusion Matrices of GPT-3.5 Predictions. The columns show the true labels as assigned by human experts, while the rows report the predictions of the system.}
  \label{tab:cm}
  \setlength{\tabcolsep}{1.9pt}
  \begin{tabular}{rrrrrrrrrrrrrrrrrrrr}
                        &\multicolumn{12}{c}{\bf CUAD}                           &             & &\multicolumn{5}{c}{\bf BVA}\\
                        &AA &AR & CS& GL & IP & IN & MC & PT & PS & TC & VR & WD &\hspace{1cm} & &CIT &EVD &FND &LR &RSN\\
Anti-assignment     &\cellcolor{black!65}\color{white}320&\cellcolor{black!0}  0&\cellcolor{black!0}  0&\cellcolor{black!0}   0&\cellcolor{black!0}   0&\cellcolor{black!0}   0&\cellcolor{black!0}   0&\cellcolor{black!0}   2&\cellcolor{black!0}   1&\cellcolor{black!0}   0&\cellcolor{black!0}   0&\cellcolor{black!0}   2&  & Citation   &\cellcolor{black!66}\color{white}1050&\cellcolor{black!1}   9&\cellcolor{black!0}   1&\cellcolor{black!1} 10&\cellcolor{black!0}  2\\
    Audit Rights        &\cellcolor{black!0}  0&\cellcolor{black!100}\color{white}489&\cellcolor{black!0}  0&\cellcolor{black!0}   0&\cellcolor{black!0}   0&\cellcolor{black!0}   2&\cellcolor{black!0}   0&\cellcolor{black!4}  18&\cellcolor{black!0}   0&\cellcolor{black!0}   0&\cellcolor{black!0}   0&\cellcolor{black!1}   4&  & Evidence   &\cellcolor{black!0}   7&\cellcolor{black!100}\color{white}1582&\cellcolor{black!2}  33&\cellcolor{black!0}  3&\cellcolor{black!5} 84\\
    Cvnt. not to Sue    &\cellcolor{black!1}  5&\cellcolor{black!1}  6&\cellcolor{black!20} 97&\cellcolor{black!0}   0&\cellcolor{black!1}   3&\cellcolor{black!1}   4&\cellcolor{black!0}   1&\cellcolor{black!3}  13&\cellcolor{black!0}   0&\cellcolor{black!0}   0&\cellcolor{black!0}   1&\cellcolor{black!0}   2&  & Finding    &\cellcolor{black!0}   0&\cellcolor{black!11} 168&\cellcolor{black!17} 274&\cellcolor{black!1} 18&\cellcolor{black!5} 87\\
    Governing Law       &\cellcolor{black!0}  1&\cellcolor{black!0}  0&\cellcolor{black!0}  0&\cellcolor{black!90}\color{white} 439&\cellcolor{black!0}   0&\cellcolor{black!0}   0&\cellcolor{black!0}   0&\cellcolor{black!0}   1&\cellcolor{black!0}   1&\cellcolor{black!0}   0&\cellcolor{black!0}   1&\cellcolor{black!0}   0&  & Legal Rule &\cellcolor{black!2}  35&\cellcolor{black!0}   3&\cellcolor{black!2}  28&\cellcolor{black!36} 572&\cellcolor{black!2} 25\\
    IP Assignment       &\cellcolor{black!3} 16&\cellcolor{black!0}  1&\cellcolor{black!4} 22&\cellcolor{black!0}   0&\cellcolor{black!49} 242&\cellcolor{black!0}   0&\cellcolor{black!0}   0&\cellcolor{black!6}  28&\cellcolor{black!2}  11&\cellcolor{black!0}   0&\cellcolor{black!0}   2&\cellcolor{black!0}   0&  & Reasoning  &\cellcolor{black!1}  20&\cellcolor{black!41} 654&\cellcolor{black!9} 149&\cellcolor{black!21} 332&\cellcolor{black!32} 504\\
    Insurance           &\cellcolor{black!0}  0&\cellcolor{black!1}  5&\cellcolor{black!0}  0&\cellcolor{black!0}   0&\cellcolor{black!0}   0&\cellcolor{black!87}\color{white} 423&\cellcolor{black!0}   0&\cellcolor{black!2}  10&\cellcolor{black!0}   0&\cellcolor{black!0}   0&\cellcolor{black!0}   0&\cellcolor{black!0}   0&\\
    Min. Commitment     &\cellcolor{black!0}  0&\cellcolor{black!1}  4&\cellcolor{black!0}  0&\cellcolor{black!0}   0&\cellcolor{black!0}   0&\cellcolor{black!0}   1&\cellcolor{black!41} 200&\cellcolor{black!2}   8&\cellcolor{black!7}  32&\cellcolor{black!0}   1&\cellcolor{black!9}  44&\cellcolor{black!1}   6&\\
    Post-term. Services &\cellcolor{black!0}  0&\cellcolor{black!2} 10&\cellcolor{black!0}  0&\cellcolor{black!0}   0&\cellcolor{black!0}   1&\cellcolor{black!0}   1&\cellcolor{black!1}   3&\cellcolor{black!29} 142&\cellcolor{black!0}   1&\cellcolor{black!1}   3&\cellcolor{black!3}  14&\cellcolor{black!1}   3&  & & &\multicolumn{3}{c}{\bf PHASYS}\\
    Profit Sharing      &\cellcolor{black!0}  0&\cellcolor{black!2}  8&\cellcolor{black!0}  0&\cellcolor{black!0}   0&\cellcolor{black!0}   1&\cellcolor{black!0}   2&\cellcolor{black!6}  30&\cellcolor{black!6}  28&\cellcolor{black!52}\color{white} 252&\cellcolor{black!0}   1&\cellcolor{black!2}   9&\cellcolor{black!0}   1& & & & PRP & RES  & REC\\
    Termination Cnv.    &\cellcolor{black!1}  7&\cellcolor{black!0}  0&\cellcolor{black!0}  1&\cellcolor{black!0}   1&\cellcolor{black!0}   0&\cellcolor{black!0}   2&\cellcolor{black!3}  16&\cellcolor{black!2}   8&\cellcolor{black!0}   0&\cellcolor{black!38} 188&\cellcolor{black!0}   0&\cellcolor{black!1}   3&  &\multicolumn{2}{r}{Preparedness}&\cellcolor{black!44} 308 &\cellcolor{black!74}\color{white} 518  &\cellcolor{black!15} 106\\
    Volume Restriction  &\cellcolor{black!1}  5&\cellcolor{black!1}  7&\cellcolor{black!0}  0&\cellcolor{black!0}   0&\cellcolor{black!0}   1&\cellcolor{black!0}   0&\cellcolor{black!8}  41&\cellcolor{black!1}   7&\cellcolor{black!2}   9&\cellcolor{black!0}   0&\cellcolor{black!13}  64&\cellcolor{black!1}   3& &\multicolumn{2}{r}{Response}&\cellcolor{black!20} 139 &\cellcolor{black!100}\color{white} 699  &\cellcolor{black!19} 134\\
    Warranty Duration   &\cellcolor{black!0}  0&\cellcolor{black!1}  3&\cellcolor{black!0}  0&\cellcolor{black!0}   0&\cellcolor{black!0}   0&\cellcolor{black!1}   6&\cellcolor{black!1}   6&\cellcolor{black!1}   5&\cellcolor{black!0}   1&\cellcolor{black!0}   0&\cellcolor{black!1}   3&\cellcolor{black!22} 109&  &\multicolumn{2}{r}{Recovery}&\cellcolor{black!0} 2   &\cellcolor{black!2}  13  &\cellcolor{black!7}  50\\
  \end{tabular}
\end{table*}

Table \ref{tab:results} shows the results of our experiments on applying the \verb|text-davinci-003| model to the three tasks involving adjudicatory opinions (BVA), contract clauses (CUAD) and statutory and regulatory provisions (PHASYS). Firstly, the GPT-3.5 model outperforms the baseline based on Jaccard similarity by a large margin on all the three tasks ($.35$ vs $.73$ on BVA, $.38$ vs $.86$ on CUAD, and $.24$ vs $.54$ on PHASYS). While the magnitude of the difference might be somewhat surprising the better performance of the GPT-3.5 model as compared to the baseline is to be expected. The baseline only matches the exact words from the type definitions to the exact words in the evaluated text snippets, whereas the sophisticated GPT-3.5 model has access to their semantics.

When compared to the supervised algorithms trained on the in-domain data, the performance of the GPT-3.5 model is surprisingly high. While the GPT-3.5 model clearly falls short when compared to the supervised models trained on large portions of the available data, it is quite competitive, perhaps beyond what could be reasonably expected, when the size of the training data is limited. Overall, it appears that the RoBERTa model needs at least several hundred training data points to match the performance of the GPT-3.5 model (light-blue cells in Table \ref{tab:results}). The random forest requires even more data, often close to a thousand (light-green cells in Table \ref{tab:results}).

The three confusion matrices in Table \ref{tab:cm} offer detailed view into the performance of the GPT-3.5 model. The performance on CUAD's contractual clauses appears very promising overall. There is a small number of classes that the system struggles to distinguish from each other, such as \emph{Minimum Commitment}, \emph{Profit Sharing}, or \emph{Volume Restrictions}. As for the BVA's adjudicatory documents, the \emph{Reasoning} class appears to be the most problematic. There is especially large number of \emph{Evidence} sentences (654) that have been misclassified as \emph{Reasoning}. Finally, the PHASYS' statutory and regulatory provisions seem to be the most challenging. A large number of \emph{Emergency Response} provisions are labeled as \emph{Emergency Preparedness}.

While the results of our experiments are promising, limitations clearly exist. First, the performance of the models is far from perfect and there is a considerable gap between the performance of the zero-shot LLM compared to the performance of the supervised ML systems trained on hundreds or thousands of example data points. Hence, in workflows with low tolerance towards inaccuracies in semantic annotation the zero-shot LLM predictions may need to be subjected to a human-expert QA. The outcome of such human-computer interaction may be a high-quality data set of the size that enables fine-tuning of a powerful domain-adapted LLM.

There are considerable differences in performance of GPT-3.5 across the three data sets. While the performance on the CUAD data set seems to be very reasonable there are some limitations when it comes to the performance on the BVA data set. The model struggles with the \emph{Reasoning} type in terms of mislabeling many sentences of other types as \emph{Reasoning} as well as not recognizing many \emph{Reasoning} sentences as such (Table \ref{tab:cm}). This is consistent with the performance of the supervised ML models. While the fine-tuned base RoBERTa is clearly more successful in handling this semantic type compared to the GPT-3.5, it still struggles (F$_1=.71$). The random forest model under-performs GPT-3.5. Hence, the correct recognition of this type may require extremely nuanced notions that may be difficult to acquire through a compact one-sentence definition (GPT-3.5) or word occurrence features (random forest). For such situations, the proposed approach might not (yet) be powerful enough and the only viable solution could be fine-tuning an LLM.

The performance of GPT-3.5 on the PHASYS data set is not satisfactory. We identified several challenges this data set poses that makes it difficult even to the supervised ML models (Table \ref{tab:results}). The data set is imbalanced with the \emph{Response} type constituting 62.4\% of the available data points. Second, the definitions of the semantic types appear to be somewhat less clear and lower quality than for the other two data sets. Hence, we hypothesize that the manual annotation of this data set heavily relied on the informal expertise of the human annotators which was not fully captured in the annotation guidelines. Finally, the fine-grained distinctions between what counts as emergency \emph{Response} as opposed to \emph{Preparedness} may simply be too nuanced to be captured in a compact definition.

\section{Conclusions}
We evaluated \verb|text-davinci-003| on three legal annotation tasks, involving adjudicatory opinions, contractual clauses, and statutory and regulatory provisions. The model was provided with a list of compact definitions of the semantic types. The tasks were to assign a batch of short text snippets with the defined categories. The results of the experiment are very promising, where the model achieved (micro) F$_1=.73$ for the rhetorical roles of sentences from adjudicatory decisions, $.86$ for the types of contractual clauses, and $.54$ for the purpose of public-health system's emergency response and preparedness statutory and regulatory provisions. Our findings are important for legal professionals, educators and scholars who intend to leverage the capabilities of state-of-the-art LLMs to lower the cost of existing high-volume workloads, involving semantic annotation of legal documents, or to unlock novel workflows that would have not been economically feasible to be performed manually or using supervised ML. We also envision that the approach could be successfully combined with high-speed similarity annotation frameworks~\cite{westermann2019computer,westermann2020sentence} to enable highly cost efficient annotation in situations where resources are scarce.

\bibliographystyle{ACM-Reference-Format}
\balance
\bibliography{main}

\end{document}